%% file: main.tex
\documentclass[sigconf,nonacm]{acmart}
\usepackage[T1]{fontenc}
\usepackage{graphicx}

\usepackage{xcolor}
\usepackage{amsmath}
\usepackage{hyperref}
\usepackage{cleveref}
\usepackage[utf8]{inputenc}
\usepackage{amsmath,amsfonts}
\usepackage{algorithmic}
\usepackage{graphicx}
\usepackage{textcomp}
\usepackage{xcolor}
\usepackage{algorithm}
\usepackage{pbox}
\usepackage{pdfpages}
\usepackage{xspace}
\usepackage{enumitem}

\usepackage{tikz}
\usetikzlibrary{positioning,decorations.markings,chains,arrows,shadows,fadings,shapes,backgrounds,snakes,matrix,patterns,plotmarks,trees,mindmap, calc, tikzmark}

\tikzset{rectstate/.style={
		draw=black,
		rectangle,
		align=center,
		minimum width=3cm
	}
}

\tikzset{edge/.style={
		->,
		draw,
		semithick,
		shorten >=1pt,
		>=stealth'
	}
}

\usepackage{listings}
\usepackage{lineno}

\newcommand{\nn}{\emph{native node}\xspace}
\newcommand{\hn}{\emph{hybrid node}\xspace}
\newcommand{\hl}{\emph{hybrid layer}\xspace}
\newcommand{\ien}{\emph{if-else node}\xspace}
\newcommand{\stf}{\emph{static feature}\xspace}
\newcommand{\df}{\emph{dynamic feature}\xspace}

\acmDOI{}
\acmISBN{}
\acmYear{2024} % Insert Publication year
\copyrightyear{2024} % Insert Copyright year (typically the same as above)
\setcopyright{cc}
\begin{document}
\title{\textbf{Register Your Forests}: Decision Tree Ensemble Optimization by Explicit CPU Register Allocation}

\author{Daniel Biebert}
\email{daniel.biebert@tu-dortmund.de}
\affiliation{%
  \institution{TU Dortmund}
  \country{Germany}
}

\author{Christian Hakert}
\email{christian.hakert@tu-dortmund.de}
\affiliation{%
  \institution{TU Dortmund}
  \country{Germany}
}

\author{Kuan-Hsun Chen}
\email{k.h.chen@utwente.nl}
\affiliation{%
  \institution{University of Twente}
  \country{Netherlands}
}

\author{Jian-Jia Chen}
\email{jian-jia.chen@cs.tu-dortmund.de}
\affiliation{%
  \institution{TU Dortmund}
  \country{Germany}
}

\renewcommand{\shortauthors}{Biebert et al.}

\begin{abstract}
    Bringing high-level machine learning models to efficient and well-suited machine implementations often invokes a bunch of tools, e.g.~code generators, compilers, and optimizers. Along such tool chains, abstractions have to be applied.
    This leads to not optimally used CPU registers.
    This is a shortcoming, especially in resource constrained embedded setups.
    In this work, we present a code generation approach
    for decision tree ensembles, which produces machine assembly code within a single conversion step directly from the high-level model representation.
    Specifically, we develop various approaches to effectively allocate registers for the inference of decision tree ensembles.
    Extensive evaluations of the proposed method are conducted in comparison to the basic realization of C code from the high-level machine learning model and succeeding compilation. 
    The results show that the performance of decision tree ensemble inference can be significantly improved (by up to $\approx1.6\times$), if the methods are applied carefully to the appropriate scenario.
\end{abstract}

\maketitle
\pagestyle{plain}
\thispagestyle{plain}
\section{Introduction}
\input{sections/intro}

\section{Related Work}
\label{sec:related}
\input{sections/related}
% \clearpage
\section{Decision Tree Models and Implementations}
\label{sec:bg}
\input{sections/dtimps}
\section{Problem Analysis}
\label{sec:problem}
\input{sections/problem}

% % \clearpage
\section{Feature Value Storing and Caching}
\label{sec:features}
\input{sections/tuplestoring}
\section{Node Storing}
\label{sec:nodes}
\input{sections/nodestoring}

\section{Evaluation}
\label{sec:eval}
\input{sections/evaluation}
\section{Conclusion and Outlook}
\label{sec:conclusion}
\input{sections/conclusion}

% \clearpage

\section*{Acknowledgement}
This paper has been supported by Deutsche Forschungsgemeinshaft (DFG), as part of the projects OneMemory (405422836), SPP 2377: Disruptive Main-Memory Technologies (460954224), and Memory Diplomat (502384507).
\bibliographystyle{ACM-Reference-Format}
\bibliography{mybibliography}

\end{document}

%% file: sections/intro.tex
Decision tree ensembles are top candidates for lightweight machine learning, especially in the field of resource constrained systems, e.g., classification in astrophysics~\cite{Buss/etal/2016a} and nanopartical analysis with biosensors~\cite{s19194138}.
Decision trees inherently provide optimization potential, since on different executions of the tree, different paths are followed. Thus, the tree inference potentially can be optimized for certain paths, e.g., several attempts have been applied to ranking models at Facebook to boost the evaluation of decision tree models\footnote{\url{https://engineering.fb.com/2017/03/27/ml-applications/evaluating-boosted-decision-trees-for-billions-of-users/}}.

Training and execution of machine learning models, so-called inference, is a data-centric task and therefore usually realized in a platform independent manner. Hence, productive programming languages as Python are popular for 
training the models~\cite{scikit-learn}. However, 
high level programming environments are likely not available in resource constrained systems, and an interpreted language also consumes considerably more energy and time~\cite{10.1145/3136014.3136031}.
In addition, recent trends show that the execution of the trained models are usually not as efficient as with a platform specific implementation. 
To bridge this gap, various solutions exist to transform the high-level models after training into platform specific models. Although these solutions are highly tuned towards performance~\cite{Dato/etal/2016, Lucchese2016, Ye/etal/2018, kim/etal/2010, Buschjaeger/Morik/2016, buschjager2018realization, 10.1145/3508019}, usually a large tool-chain needs to be employed.
However, within the tool-chain, the output of the different tools need to be sufficiently abstract and generic to serve as inputs for next tools. Such abstractions also induce possible information loss, causing a diminished potential for architectural optimization.

In this work, we investigate if a direct implementation generation with fewer abstraction layers is capable to optimize the execution performance of decision tree ensembles. With a certain amount of explicitly allocatable hardware registers, our objective is to manage the register allocation in the implementation to achieve a performance improvement, by accelerating multiple accesses to the same value.
Towards that, we distinguish native trees, in which each tree is an array-based implementation of tree nodes, and if-else trees, in which each tree is a representation of tree nodes with nested if-else statements. The two variants put stress on different parts of the CPU. A native tree loads all tree data from data memory and therefore utilizes data caches, while an if-else tree loads all values from instruction memory, employing instruction caches. We further propose a third alternative, implementing a part of the tree as an if-else tree and executing the rest of the tree as a native tree. This approach could make use of the different advantages of both variants while leveling the load over the different caches.

\noindent\textbf{Our contributions:}
\begin{itemize}
    \item An implementation of decision trees to store node values in CPU registers for X86 and ARMv8
    \item Strategies for choosing which node values to store in registers for native trees and if-else trees and a hybrid combination of both
    \item Extensive experimental evaluation of the performance gain on a large set of ensembles and data sets
\end{itemize}

%% file: sections/related.tex
Although training and execution of decision trees are famous on a abstract level, related work investigates the possibilities to implement decision trees on hardware. Van Essen et al. compare different ad-hoc realizations across varying computing architectures, including CPUs, GPUs and FPGAs \cite{VanEssen/etal/2012}.  Nakandaka et al. bring decision trees even more specific to parallel execution architectures, by converting trees into tensor operations and exploit massive parallelism \cite{nakandala2020tensor}. As a generic, hence optimized CPU based implementation, Asadi et al.~introduce the popular approaches of native and if-else tree implementations \cite{Asadi/etal/2014}. 
Dato et al.~optimize decision tree inference by keeping the models itself untouched, but reorder the execution sequence in the implementation \cite{Dato/etal/2016}.
Lucchese et al.~further include vectorization into the implementation in order to accelerate parallel execution of multiple examples \cite{Lucchese2016}.
Different concepts in order to accelerate machine-specific execution are further proposed.
Ye et al.~introduce run length encoding for the tree execution in order to compact the data representation \cite{Ye/etal/2018}. Kim et al.~propose a tree implementation, which is optimized to the memory hierarchy on modern CPUs and GPUs \cite{kim/etal/2010}.
Buschjaeger and Morik consider different realization schemes for tree traversal and theoretically analyze their execution time, based on a probabilistic view~\cite{Buschjaeger/Morik/2016}.

Considering a broader scope beyond decision tree ensembles, C / C++ generators for other machine learning models (e.g.~neural networks) have emerged recently \cite{david2020tensorflow, warden2019tinyml}. Architecture-aware software designs also demonstrate their great benefit to speed up the inference of other machine learning models in the literature, e.g., most for various neural networks~\cite{9499860, 8416862, 9499917, 10.1145/3297858.3304011, 10.1145/3297858.3304076}, and some for random forests~\cite{buschjager2018realization, 10.1145/3508019}.
Although these architecture-aware approaches exist, to the best of our knowledge, no method exists, which transforms random forests to platform-specific assembly code by exploiting hardware features and performing explicit register allocation. Khorasani et~al. perform explicit register management in GPUs \cite{khorasani2018register}, which is a different approach to explicit CPU register allocation.

%% file: sections/dtimps.tex
Generally, two implementation types can be distinguished for decision trees: 1) native trees, where tree nodes become data objects and a narrow loop iterates over them and 2) if-else trees, where tree nodes become nested if-else blocks and the tree is visited by directly jumping into the if or the else block \cite{Asadi/etal/2014}. In this section, we lay out a logical model of a decision tree first and afterwards describe how both implementations can be derived.

\subsection{Probabilistic Model}
We assume decision tree $X$ consist of nodes $\{n_0, n_1, ..., n_{m-1}\}$, where $n_0$ is the root node. Each node has a feature index $FI(n_i)$, a split value $S(n_i)$ and a left and right child index $LC(n_i), RC(n_i)$. The tree is then visited in such a way, that subsequently after node $i$, node\\
$n_j=\left\{
\begin{array}{c|c}
     LC(n_i)&input\_data[FI(n_i)] \leq S(n_i)\\
     RC(n_i)&input\_data[FI(n_i)] > S(n_i)
\end{array}\right.$ is accessed, where $input\_data$ is the feature vector to be inferred. Once a leaf node is reached, it contains a prediction value $P(n_i)$, which forms the output of the decision tree.

During training, how many data tuples leading to the left or the right child can be traced, resulting in a relative probability of the left and right child node. We denote by $prob(n_i)$ the probability of node $n_i$ to be accessed from the parent node. Naturally, $prob(LC(n_i))+prob(RC(n_i))=100\%$ and $prob(n_0)=100\%$. Every node in the tree has a unique access path $path(n_i)=\{pn_0, pn_1, ..., pn_j\}$, where $pn_0=n_0, pn_j=n_i$ and $pn_{k+1}=LC(pn_k)$ or $pn_{k+1}=RC(pn_k)$. With the help of this path, we can define an absolute node probability $absprob(n_i)$:
\begin{equation}
    absprob(n_i)=\Pi_{pn_j \in path(n_i)} prob(pn_j)
\end{equation}
This allows to rank nodes of the decision tree according to their probability to be accesses in the training data-set. We expect a similar distribution of accesses in a productive test data-set, thus when optimization is applied according to this node ranking, it is expected to also optimize the tree for test data.

\subsection{Tree Implementation}
In the realization of native trees every node is stored as a tuple of all its element, namely the feature index, the split value and the left and right indices. These values are stored in a data array. The inference is performed by a small loop over the data array. Starting with the root of the tree as the current node, the feature value is compared to the split value and depending on the outcome the next current node will be the left or the right child. This is repeated until a leaf is reached. At that point the prediction is returned.
This tree realization intensively uses data memory and only allocates a small footprint of instruction memory. The code, however, is also fixed for all nodes, which makes it harder to modify single nodes.

In the realization of if-else trees every node becomes a single piece of code. The nodes then become nested if-else blocks in the inference code. All values are used as immediates and therefore also stored in instruction memory. Although this has the clear advantage to utilize fewer data memory and to provide modified source code for single nodes, the order of nodes in memory cannot be arbitrarily changed. Child nodes of a certain node must always appear in the if or in the else branch. Within this realization, feature and split and prediction are constants, which are assigned at compile time and are therefore immediately present in the source code.

Both approaches can also be combined, where a set of nodes in an if-else tree can be realized as a native tree. Within one ensemble, trees can be arbitrarily implemented as native trees, if-else trees or a combination of both. 
To the best of our knowledge, such hybrid realizations have not been explored in the literature.

%% file: sections/problem.tex
In this work, we present novel implementations of decision trees, where we utilize available registers to realize a custom caching of decision tree contents within the CPU registers. 
Usually, such registers are managed by the compiler, following ABI conventions. The execution of tree inference, however, can be realized as a single function in C / C++. The registers, which are not written or read by compiler-generated code, can be reserved for arbitrary usage via inline assembly within a function. We explicitly use them in the generated custom code.

\subsection{System Model}
We assume a generic structure of registers, where a system has a number $\#R$ of registers. To adapt to general technical limitations, we assume $r < \#R$ of them can be arbitrarily used. We assume that we can load memory content into each of these registers with a corresponding load instruction and can store contents to memory. We further assume that we can perform the required arithmetic operations on the registers, which are required for decision tree inference.
This includes comparison of a register value with another register or with a memory content and succeeding conditional branches. Real systems usually provide different types of registers, e.g.~general purpose registers (GPR), floating point registers (FLP), vector registers, etc. By keeping track of the register type, corresponding assembly code can be generated to perform the aforementioned operations on all types of registers.
Since we explicitly allocate registers during inference, the developed implementations are also broadly compatible with parallelization on the granularity of trees.
\subsection{Problem Definition}
Given a trained ensemble, consisting of decision trees with profiled probabilities on the training data set and an amount $r$ of arbitrarily allocatable registers, the problem is to realize an implementation, which stores tree contents in the registers without altering the logic tree structure 
Data tuples are processed in a batch manner for each decision tree, by which certain nodes are likely accessed multiple times.
If values are repeatedly accessed, a value from a register can be faster accessed compared to a value from memory. On the downside, a value in a register blocks this register for other usage and therefore potentially can slow down the entire program.

%% file: sections/tuplestoring.tex
As one approach towards solving the defined problem, we first investigate how to handle the temporary input data with custom register allocation.
In each run of the ensemble, one input tuple is classified (for inference) utilizing the decision trees. Depending on the size of the trees and the number of tuples, some values might be accessed multiple times while following the trees on one path. 

We propose two methods to store the most suitable feature values in registers. First, a static approach (referenced with \stf), where the feature values of a native tree are loaded into registers before utilizing the tree and reside in the registers for the duration of following a path in each tree of the ensemble. Secondly, a dynamic approach (referred to as \df), in which the feature values are dynamically allocated to registers in an if-else tree.

\subsection{Selection Policy}
\label{sec:selectedValue}
Input tuples potentially have high dimensions, where certain entries may be accessed more frequently than others. Thus, we propose a metric to select a subset of the tuple entries to be forced to registers.

\begin{equation} \label{eq:suitability}
    S_i = \sum_{j=1}^{depth(tree)} p_{i,j} * j
\end{equation}

\Cref{eq:suitability} determines the suitability of a feature value at index $i$. $p_{i,j}$ is the probability of the feature value index $i$ to be accessed $j$ times. These probabilities are then added for every possible $j$, being weighted by $j$ itself. This ensures that the probability of a feature value being used more often during one single inference is weighted higher. We then pick the $n$ highest scored values for register storing.  These values reside in the register for the duration of the inference within one function call.
To check whether a feature value is in a register or in memory, the index of the to access feature value can be compared. The feature value is either loaded from memory or directly used from a register.

\subsection{Dynamic Feature Caching}

Alternatively, we develop an approach that allows registers to cache feature values dynamically. This can be utilized in an if-else tree without significant efforts, because every node in an if-else tree receives its own distinct code. Suitable access code to a certain register can be directly integrated. To fully use this attribute, we dynamically cache the feature values inside the registers during run time. The feature values are loaded into registers when being used first. If it is already in a register, it can be used directly from the register itself. In case there are no spare registers for storing, the oldest cached value is replaced accordingly.
Although this type of register management introduces large overheads, they are all present during code generation. Once the tree is constructed, the register caching allocation can be statically defined, and the corresponding code is directly generated into the if-else blocks.

%% file: sections/nodestoring.tex
Besides storing feature values in registers, we investigate an alternative approach where nodes of the tree data structure are stored in registers. 
To select the nodes to be stored in registers, we order tree nodes along their absolute access probability $absprob(n_i)$ and select the first $r$ nodes.
To improve the performance, the part of the tree loaded into registers is kept in registers when the tree is visited multiple times.

\subsection{Realization of Register Cached Nodes}

Implementing the whole tree as a native tree and accessing the correct values using a comparison chain is the first version, proposed in this paper (\nn): Every time a new node is accessed, a comparison with the current index is performed to determine the location of the node values. These are then either loaded from the corresponding register or loaded from memory.
We further proposed another approach where the nodes picked for register storage are implemented and visited as nodes from an if-else subtree (we refer to such nodes as if-else nodes) within the native tree.
Here, two different subsets of the tree can be implemented as if-else nodes. Firstly, implementing the first layer nodes (\hl) and secondly implementing the most probable nodes (\hn) as if-else nodes.
Ultimately, the full tree can be implemented as an if-else tree with the most probable nodes stored in registers. We call this approach the \ien method. Consequently, the data in the nodes, which are stored in a register, is used directly, omitting the need to load this value from instruction memory.

\subsection{Tree Construction and Inference}
Registers have limited space, so not all node values should be stored, but only the ones necessary for inference.
The most important value in a node is the split value. It is used to compare with the feature value. As such, it is used in every node except leaf nodes.
If the split value does not fill the register completely, more values can be stored. Three values are of key interest here: left, right child indices and the feature index. Depending on the size of the register, not all can be stored at once.
Different realizations are possible.
If the node is executed as part of a native subtree (referenced by native node), the indices for the left and right children are required to load the new node. Therefore, left and right child indices are loaded into the register.
In some situations, left and right child indices are not needed. If the children are implemented as an if-else nodes, the indices of the children are superfluous.
In these situations, the rest of the register is filled with the feature index.
When implementing the \hn method, some nodes may have one child as an if-else node and another one as a native node. In these situations, the index of the child implemented as a native node and the feature index can both be loaded into the the register.
To access the nodes in registers, different implementations are required for native nodes and if-else nodes. In native trees, the individual nodes are identified by their index. To access the node data inside registers, comparisons with the index of the current node determines, where to load the data from. No further implementation is needed for if-else nodes, as each node receives separate code accessing the node data.

%% file: sections/evaluation.tex
\begin{figure}
    \centering
    \caption{Native tree methods on X86 for server class (top) and desktop class (down) - 100 Trees with max. depth 15}
    \includegraphics[width=\linewidth]{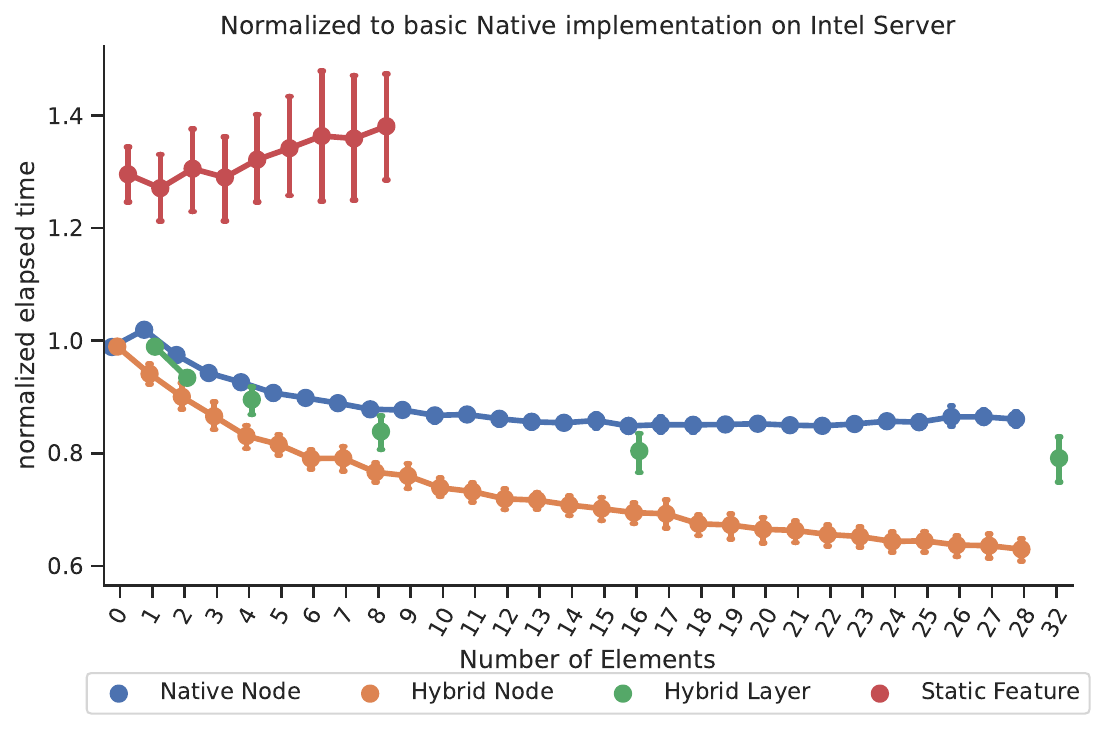}
    \includegraphics[width=\linewidth]{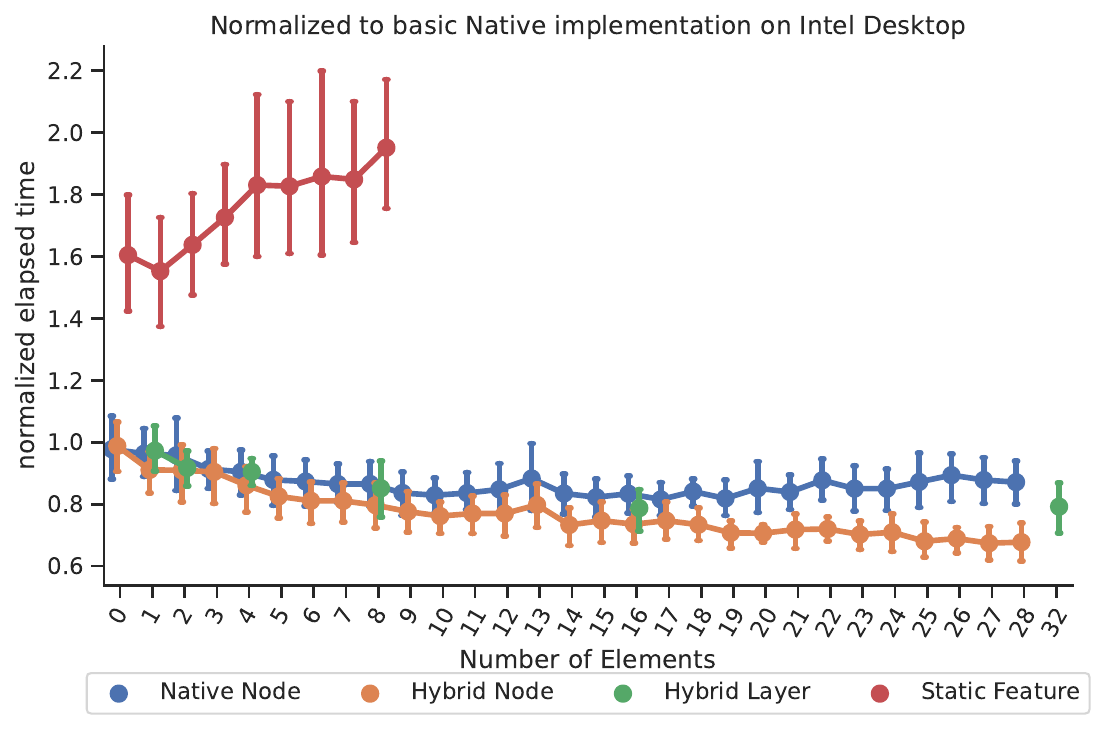}
    \label{fig:nativex86}
    \vspace{-1em}
\end{figure}

To assess the improvement in terms of execution time of the proposed methods, extensive tests are run on different hardware and multiple kinds of trees: 9 data-sets under 6 proposed methods. The data-sets (adult, bank, covertype, letter, magic, satlog, sensorless-drive, spambase and wine-quality) are taken from the UCI Machine Learning Repository \cite{Dua:2019} and trained using sklearn.
Analyzing the effect the depth of the tree can have on the performance, experiments are run with a single decision tree and a random forest with 25 trees, using a maximum depth of 5, 10, 15 and 20. Further, the experiments are also run with a random forest with maximum depth 15 and using different numbers of trees: 10, 25, 50 and 100.
To serve as a baseline, all configurations are run once with a basic C++ implementation of both a native tree and an if-else tree.
All experiments are compiled with gcc on default settings.

These experiments are run on both server and desktop class CPUs using the X86 and ARMv8 architecture. For the X86 machines, we use an \textbf{AMD EPYC 7742} system (256GB DDR4 RAM) for the server and an \textbf{Intel Core I7-8550U} system (16GB DDR4 RAM) for the desktop. For the ARMv8 machines, we use a \textbf{Cavium Thunder X2} system (256GB DDR4 RAM) for the server and an \textbf{Apple M1} system (16GB DDR4 RAM) for the desktop. The X86 systems feature 16 general purpose registers with 64 bit and 16 floating point registers with 128 bit, while the ARMv8 systems feature 32 general purpose 64 bit registers and 32 64 bit floating point registers, respectively.

\subsection{Native Tree Modifications}

Our proposed methods can be roughly categorized into methods modifying native trees (\hn(HN), \hl(HL), \nn(NN)) and into methods modifying if-else trees (\ien(IN) and \df(DF)).
Please note that the \hl and \hn methods implement native trees for the major part, though they introduce partial if-else trees, resulting in hybrid trees. 

Focusing on the native tree methods, \Cref{fig:nativex86} depicts normalized results to the basic native tree implementation on the X86 machine.
We run all experiments in a row and normalize the execution time of the various methods to the basic baseline. We print the number of elements allocated in registers on the x axis and the normalized execution time (larger than 1 means slower than the baseline) on the y axis. We consider all data-sets and print the average as a point together with the variance. Every plot depicts a concrete ensemble configuration with a fixed number of trees and a limited maximal height of the trees. Note that for the \hl method registers can be only allocated in numbers of powers of 2, since entire tree layers are allocated to registers.
The figures illustrate only one particular ensemble configuration for a maximum tree depth of 15 and an ensemble of 100 trees on the X86 machine and 50 on the ARMv8 machine, where effects of the different methods can be observed. Effects on the execution time across all ensemble sizes can be observed in the reported geometric means of normalized execution time for both all depths and small trees (limited depth up to $10$) and large trees (limited depth above $10$) separately (\Cref{table:geometric}).

\newcommand{\pcolor}{green!60!black}
\newcommand{\ncolor}{red!60!black}
\begin{table}
    \centering
    \caption{Geometric mean of normalized execution time, compared to native and if-else for small ($D\leq 10$) and large ($D > 10$) trees (-S: Server, -D: Desktop, s: small, l: large)}
    \label{table:geometric}
    \begin{tabular}{c|cccccccc} 
         \textbf{Method}&\multicolumn{2}{c}{X86-S}&\multicolumn{2}{c}{X86-D}&\multicolumn{2}{c}{ARM-v8-S}&\multicolumn{2}{c}{ARM-v8-D}\\\hline
         &\multicolumn{8}{c}{\textbf{native baseline}}\\\hline
        %  &X86-S&X86-D&ARM-v8-S&ARM-v8-D\\\hline
        \textbf{NN}&\multicolumn{2}{c}{\textcolor{\pcolor}{$0.86$}}&\multicolumn{2}{c}{\textcolor{\pcolor}{$0.87$}}&\multicolumn{2}{c}{\textcolor{\ncolor}{$1.29$}}&\multicolumn{2}{c}{\textcolor{\ncolor}{$1.00$}}\\
         \textbf{ST}&\multicolumn{2}{c}{\textcolor{\ncolor}{$1.04$}}&\multicolumn{2}{c}{\textcolor{\ncolor}{$1.19$}}&\multicolumn{2}{c}{\textcolor{\ncolor}{$1.22$}}&\multicolumn{2}{c}{\textcolor{\pcolor}{$0.88$}}\\
         \textbf{HN}&\multicolumn{2}{c}{\textcolor{\pcolor}{0.71}}&\multicolumn{2}{c}{\textcolor{\pcolor}{$0.71$}}&\multicolumn{2}{c}{\textcolor{\pcolor}{$0.74$}}&\multicolumn{2}{c}{\textcolor{\pcolor}{$0.62$}}\\
         \textbf{HL}&\multicolumn{2}{c}{\textcolor{\pcolor}{0.74}}&\multicolumn{2}{c}{\textcolor{\pcolor}{$0.74$}}&\multicolumn{2}{c}{\textcolor{\pcolor}{$0.82$}}&\multicolumn{2}{c}{\textcolor{\pcolor}{$0.66$}}\\\hline
         
         &s&l&s&l&s&l&s&l\\\hline
         
         \textbf{NN}&\textcolor{\pcolor}{$0.86$}&\textcolor{\pcolor}{$0.85$}&\textcolor{\ncolor}{$1.32$}&\textcolor{\ncolor}{$1.01$}&\textcolor{\pcolor}{$0.86$}&\textcolor{\pcolor}{$0.88$}&\textcolor{\ncolor}{$1.27$}&\textcolor{\pcolor}{$9.99$}\\
         \textbf{SF}&\textcolor{\pcolor}{$0.99$}&\textcolor{\ncolor}{$1.03$}&\textcolor{\ncolor}{$1.07$}&\textcolor{\pcolor}{$0.82$}&\textcolor{\ncolor}{$1.07$}&\textcolor{\ncolor}{$1.30$}&\textcolor{\ncolor}{$1.32$}&\textcolor{\pcolor}{$0.92$}\\
         \textbf{HN}&\textcolor{\pcolor}{0.66}&\textcolor{\pcolor}{$0.62$}&\textcolor{\pcolor}{$0.63$}&\textcolor{\pcolor}{$0.54$}&\textcolor{\pcolor}{0.75}&\textcolor{\pcolor}{$0.77$}&\textcolor{\pcolor}{$0.81$}&\textcolor{\pcolor}{$0.66$}\\
         \textbf{HL}&\textcolor{\pcolor}{0.68}&\textcolor{\pcolor}{$0.66$}&\textcolor{\pcolor}{$0.72$}&\textcolor{\pcolor}{$0.57$}&\textcolor{\pcolor}{0.78}&\textcolor{\pcolor}{$0.79$}&\textcolor{\pcolor}{$0.88$}&\textcolor{\pcolor}{$0.72$}\\\hline\hline
         
        &\multicolumn{8}{c}{\textbf{if-else baseline}}\\\hline
        %  &X86-S&X86-D&ARM-v8-S&ARM-v8-D\\\hline
         \textbf{IN}&\multicolumn{2}{c}{\textcolor{\pcolor}{$0.95$}}&\multicolumn{2}{c}{\textcolor{\pcolor}{$0.96$}}&\multicolumn{2}{c}{\textcolor{\ncolor}{$1.06$}}&\multicolumn{2}{c}{\textcolor{\pcolor}{$0.99$}}\\
         \textbf{DF}&\multicolumn{2}{c}{\textcolor{\pcolor}{$0.91$}}&\multicolumn{2}{c}{\textcolor{\pcolor}{$0.92$}}&\multicolumn{2}{c}{\textcolor{\ncolor}{$1.04$}}&\multicolumn{2}{c}{\textcolor{\pcolor}{$0.88$}}\\
         \textbf{HN}&\multicolumn{2}{c}{\textcolor{\ncolor}{1.77}}&\multicolumn{2}{c}{\textcolor{\ncolor}{$2.00$}}&\multicolumn{2}{c}{\textcolor{\ncolor}{$1.66$}}&\multicolumn{2}{c}{\textcolor{\ncolor}{$1.75$}}\\
         \textbf{HL}&\multicolumn{2}{c}{\textcolor{\ncolor}{1.84}}&\multicolumn{2}{c}{\textcolor{\ncolor}{$2.08$}}&\multicolumn{2}{c}{\textcolor{\ncolor}{$1.84$}}&\multicolumn{2}{c}{\textcolor{\ncolor}{$1.87$}}\\\hline
         
         &s&l&s&l&s&l&s&l\\\hline
         
         \textbf{IN}&\textcolor{\pcolor}{$0.94$}&\textcolor{\pcolor}{$0.96$}&\textcolor{\ncolor}{$1.06$}&\textcolor{\pcolor}{$0.93$}&\textcolor{\pcolor}{$0.95$}&\textcolor{\pcolor}{$0.95$}&\textcolor{\ncolor}{$1.06$}&\textcolor{\ncolor}{$1.03$}\\
         \textbf{DF}&\textcolor{\pcolor}{$0.92$}&\textcolor{\pcolor}{$0.94$}&\textcolor{\ncolor}{$1.07$}&\textcolor{\pcolor}{$0.81$}&\textcolor{\pcolor}{$0.90$}&\textcolor{\pcolor}{$0.91$}&\textcolor{\ncolor}{$1.02$}&\textcolor{\pcolor}{$0.92$}\\
         \textbf{HN}&\textcolor{\ncolor}{1.47}&\textcolor{\ncolor}{$1.65$}&\textcolor{\ncolor}{$1.43$}&\textcolor{\ncolor}{$1.36$}&\textcolor{\ncolor}{1.97}&\textcolor{\ncolor}{$2.23$}&\textcolor{\ncolor}{$1.81$}&\textcolor{\ncolor}{$2.03$}\\
         \textbf{HL}&\textcolor{\ncolor}{1.52}&\textcolor{\ncolor}{$1.75$}&\textcolor{\ncolor}{$1.63$}&\textcolor{\ncolor}{$1.41$}&\textcolor{\ncolor}{2.06}&\textcolor{\ncolor}{$2.29$}&\textcolor{\ncolor}{$1.97$}&\textcolor{\ncolor}{$2.20$}\\

    \end{tabular}
    \vspace{-1em}
\end{table}

It can be observed that the \stf method cannot improve the performance in most configurations among the basic native tree. The overheads, introduced to distinguish if a feature value should be loaded from a register or memory overshoots the improvement in the access latency. The \nn method can be observed to behave different on both CPU architectures. On X86, this method can reduce the total execution time to $\approx 85\%$ of the basic native tree, which makes a speedup of $\approx 1.17\times$. On the ARMv8 server contrarily, this method diminishes the performance overall while it also features no significant improvement on the ARMv8 desktop. This suggests the conclusion that the efficiency of the comparison chain is architecture dependent and thus, this method should be considered depending on the architecture. These trends are also consistent with considering the geometric mean across ensembles sizes, data-sets, register sizes $r \geq 5$ and storing full nodes or only split values. These values can be found in \cref{table:geometric}. The \stf reports an average increase of $1.11\times$ on X86, an increase of $1.22\times$ on the ARMv8 server and actually decreases to $0.88\times$ on the ARMv8 desktop. Examining the \nn method, the explained results of a measurable performance improvement on X86 can be seen. On the ARMv8 desktop, the performance stays roughly the same.

The \hl and \hn methods both achieve significant execution time improvements on both machines for a sufficient number of available registers, even up to a reduction to $\approx 60\%$ of the basic tree, which is a speedup of $\approx 1.66\times$.
As before \Cref{table:geometric} shows the explained geometric mean across all data-sets.
A significant improvement of the execution time can be observed for all considered systems. According to \Cref{table:geometric} it can be further observed, that most methods perform better on smaller trees than on larger trees. This can be caused by the fact that more values can reside in registers during repetitive inference on smaller trees.

Investigating these results, we report that careful attention has to be put to the setup, when explicit register management is used to improve the performance of native tree ensembles. Not only a wrong approach can diminish the overall performance, depending on the CPU architecture various approaches can result in different speedups. However, insights from this evaluation can be used to chose appropriate methods for a certain CPU architecture (e.g. the \hn method for a large amount of allocated registers) and reduce the total execution time for inference down to $\approx 58\%$ of the basic method. It can also be reported that this improvement can be gained for a sufficient number of available registers and is not linearly scaling down for more available registers.

\subsection{If-Else Tree Modifications}
\begin{figure}
    \centering
    \caption{If-else tree methods on X86 for server class (top) and desktop class (down) - 25 Trees with maximum depth 5}
    \includegraphics[width=\linewidth]{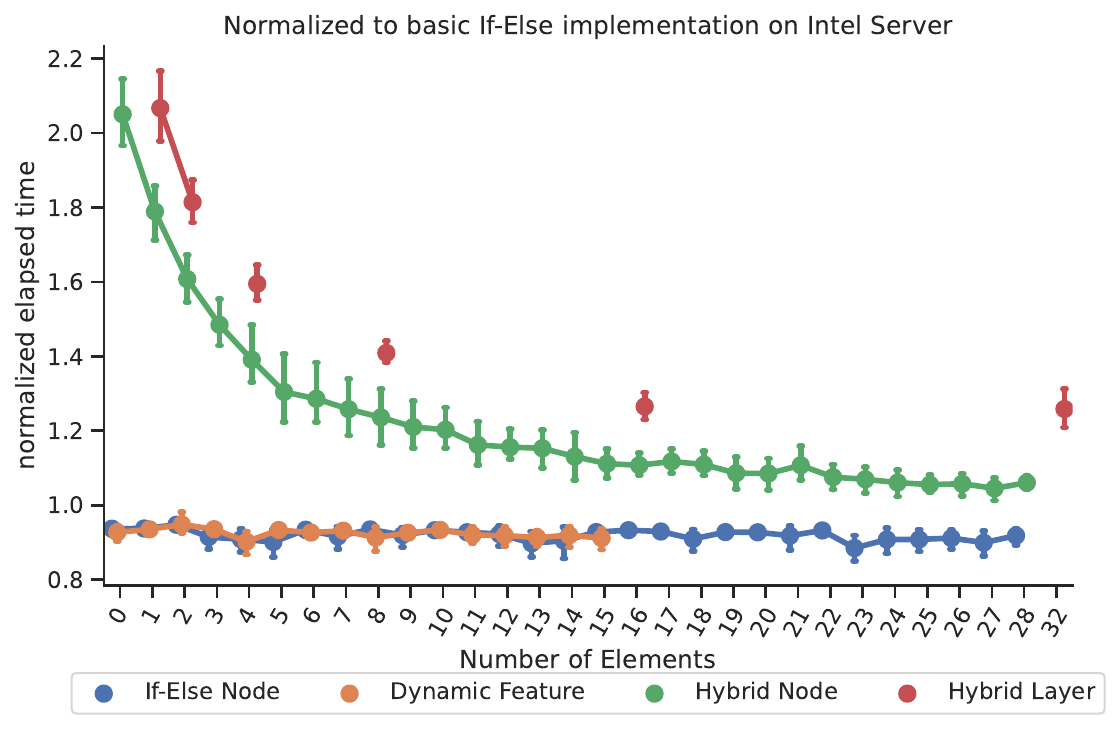}
    \includegraphics[width=\linewidth]{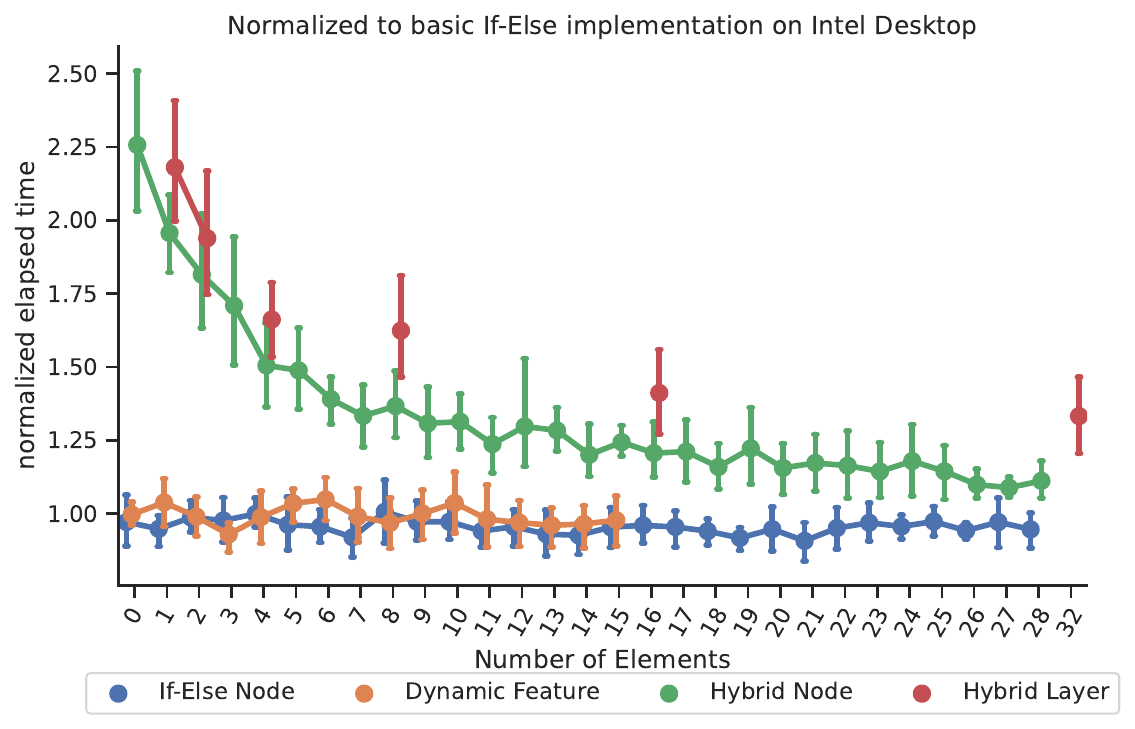}
    \vspace{-2em}
\end{figure}
The proposed methods implementing if-else trees are \df, where nodes are stored in registers during inference in a cache wise manner and \ien, with nodes stored in registers statically.
\Cref{fig:ielse86} illustrates the same ensemble size as before for the if-else tree methods. Thus, the reported numbers are also normalized to the basic if-else tree implementation. We further investigate the \hn and \hl method in comparison to basic if-else trees, since these methods partially implement if-else trees.

Investigating this case, highlights that the performance of the \hn method introduces larger overheads in comparison to a native if-else tree, which cannot be leveraged by the performance benefit. The same trend can be observed for the \hl method.
Consequently, the \hn and \hl method offer a trade-off between native trees and if-else trees, but cannot outperform if-else trees.
The \ien and \df method can improve the performance down to $70\%$ of the basic if-else tree, but only for a particular system and ensemble configuration.

Considering the geometric means across all configurations and datasets (\Cref{table:geometric}), on X86, it reports to a decrease of $0.86\times$ for the if-else tree registers method on the server and $0.87\times$ on the desktop and of $0.91\times$ (server) and $0.91\times$ (desktop) for the \df method, compared to the basic if-else tree. On the ARMv8 server, the increase reports as $1.06\times$ for the \ien method and $1.04\times$ for the \df method, respectively. On the ARMv8 desktop, performance is decreased to $0.99\times$ for the \ien method and $0.88\times$ for the \df method. The \hn and \hl methods report a consistent increase of execution time. These results conclude that, explicit register allocation can significantly improve performance, but it is highly dependent on the system architecture and the ensemble configuration. If a wrong method is deployed, performance can be drastically decreased.

Overall, for the considered data sets, CPU architectures and ensemble configurations, the \hn method achieves the best performance for native trees, while the \df method and the \ien method achieve the best performance for if-else trees. The results further suggest choosing these methods with a sufficient amount of registers. 
For the native trees, the results show a number of $\approx 20$ registers to be a good choice, while for if-else trees fewer registers ($\approx 10$) report to be sufficient.

%% file: sections/conclusion.tex
In this paper, we present an approach to generic and platform independent tool-chains for decision tree ensemble implementations, where we directly generate machine specific assembly code from the high-level model representation. Thereby, we explicitly take over the control of the allocation of CPU registers and allocate certain decision tree nodes permanently to a register. We do this for both, native trees and if-else trees.
Experimental evaluation clearly points out that explicit register allocation does not unconditionally improve the performance of decision tree inference, but with careful attention to the scenario, significant performance improvement can be achieved. If the right method is applied to the right scenario, the execution times of native trees can be decreased down to $0.58\times$ and the execution time of if-else trees down to $0.7\times$, respectively.

For future work, we plan to expand the scope of explicit register allocation to the entire ensemble. Scheduling the execution of multiple trees may open a huge design space for further optimization.